\documentclass[letterpaper]{article} 
\usepackage{aaai24}  
\usepackage{times}  
\usepackage{helvet}  
\usepackage{courier}  
\usepackage[hyphens]{url}  
\usepackage{graphicx} 
\usepackage{natbib}  
\usepackage{caption} 
\usepackage{algorithm}
\usepackage{algorithmic}
\usepackage{amsmath,amsfonts}
\usepackage{tabularx}
\usepackage{tabulary}
\usepackage{multirow}
\usepackage{makecell}
\usepackage{algorithm,algorithmic,paralist}
\usepackage[table, dvipsnames]{xcolor}

\usepackage{array,booktabs}
\usepackage{newfloat}
\usepackage{listings}
\DeclareCaptionStyle{ruled}{labelfont=normalfont,labelsep=colon,strut=off} 
\floatstyle{ruled}
\newfloat{listing}{tb}{lst}{}
\floatname{listing}{Listing}
\title{Entropic Open-set Active Learning}
\author {
    Bardia Safaei\textsuperscript{\rm 1},
    Vibashan VS \textsuperscript{\rm 1},
    Celso M. de Melo\textsuperscript{\rm 2},
    Vishal M. Patel\textsuperscript{\rm 1}
}
\affiliations {
    \textsuperscript{\rm 1}Johns Hopkins University, Baltimore, MD, USA\\
    \textsuperscript{\rm 2}DEVCOM Army Research Laboratory, Adelphi, MD, USA\\
    \{bsafaei1, vvishnu2\}@jhu.edu, 
    celso.m.demelo.civ@army.mil,
    vpatel36@jhu.edu
}

\usepackage{bibentry}

\begin{document}

\maketitle

\begin{abstract}
Active Learning (AL) aims to enhance the performance of deep models by selecting the most informative samples for annotation from a pool of unlabeled data. Despite impressive performance in closed-set settings, most AL methods fail in real-world scenarios where the unlabeled data contains unknown categories. Recently, a few studies have attempted to tackle the AL problem for the open-set setting. However, these methods focus more on selecting known samples and do not efficiently utilize unknown samples obtained during AL rounds. In this work, we propose an Entropic Open-set AL (EOAL) framework which leverages both known and unknown distributions effectively to select informative samples during AL rounds. Specifically, our approach employs two different entropy scores. One measures the uncertainty of a sample with respect to the known-class distributions. The other measures the uncertainty of the sample with respect to the unknown-class distributions. By utilizing these two entropy scores we effectively separate the known and unknown samples from the unlabeled data resulting in better sampling. Through extensive experiments, we show that the proposed method outperforms existing state-of-the-art methods on CIFAR-10, CIFAR-100, and TinyImageNet datasets. Code is available at \url{https://github.com/bardisafa/EOAL}.

\end{abstract}
\section{Introduction}

In recent years, deep learning methods have shown remarkable performance in a large number of complex computer vision tasks such as classification \cite{he2016deep, radford2021learning}, segmentation\cite{chen2017deeplab, kirillov2023segment} and object detection \cite{ren2015faster,redmon2016you}. However, the success of these deep learning models in solving these complex tasks heavily relies on the availability of extensive labeled data \cite{vs2023mask, vs2022mixture}. Obtaining labeled data is generally labor-intensive, and expensive \cite{wei2015submodularity, vs2023towards}.  Active Learning (AL) tackles this huge data labeling issue by strategically selecting a subset of informative samples for annotation, rather than labeling the entire data. Primarily, there are two types of AL techniques: a) uncertainty-based methods, and b) diversity-based methods. Uncertainty-based techniques \cite{seung1992query} leverage model uncertainty on unlabeled samples to select the most informative ones, while diversity-based methods \cite{nguyen2004active} focus on enhancing model learning by carefully choosing samples that show maximal diversity.

In general, AL methods produce promising results in closed-set settings where the unlabeled data contains only known classes. However, in real-world scenarios, this assumption does not hold as the unlabeled data contains both known and unknown samples. As a result, the performance of these closed-set AL methods significantly declines \cite{ning2022active}.  One main reason for this phenomenon is that existing uncertainty- and diversity-based methods choose the unknown samples as the most informative samples for human annotation, thereby wasting the annotation budget. Human annotators would disregard these unknown samples because they are unnecessary for the target task. Therefore, it is important to address the problem of active learning under an open-set scenario where unknown class samples might appear in the unlabeled data during sampling. 
\begin{figure}[t!]
    \begin{center}
        \includegraphics[width=1.0\linewidth]{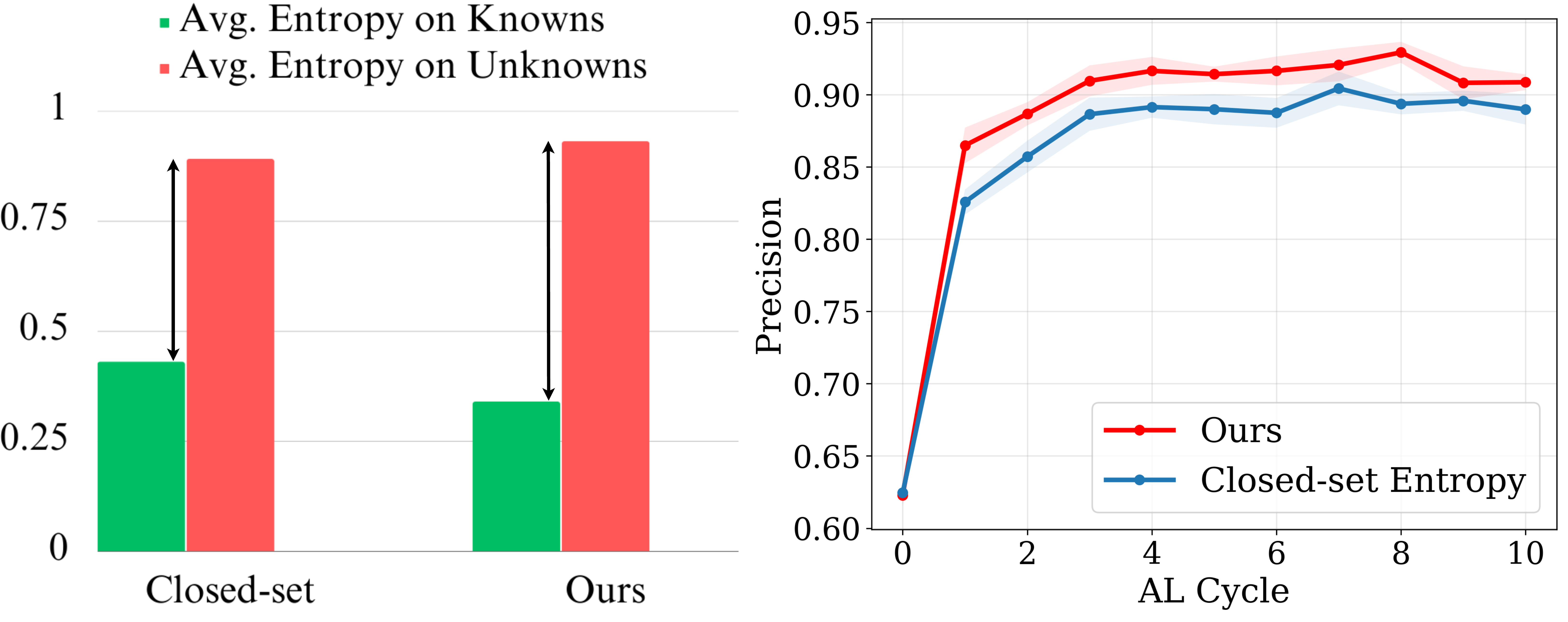}
    \end{center}
    \caption{
    \textbf{Left}: At the $5$-th AL round, we plot the difference between the average entropy scores of known and unknown samples using a closed-set classifier and our method. Our method utilizes two entropy scores which enhance the separation between known and unknown samples resulting in a better sampling of the knowns.  \textbf{Right}: Active sampling precision graph for a closed-set classifier and our method. Because of better separation between known and unknown samples, our method tends to have better precision in sampling known samples at each AL cycle. (CIFAR-10, $40\%$ mismatch ratio)
    }
    \label{fig:motivation} 
\end{figure}

A straightforward open-set AL approach is to train a closed-set classifier and utilize entropy scores to separate known and unknown samples. Later, we annotate the low entropy samples because the closed-set classifier produces low entropy for known samples (see Fig. \ref{fig:motivation}).
However, acquiring unknown samples together with know-class samples is unavoidable, especially when their presence is extensive. To address this challenge, LfOSA \cite{ning2022active} proposes to utilize unknown samples and train a classifier to reject the unknown samples and focus more on selecting known samples. MQNet \cite{park2022meta} proposes a more agnostic approach where the model leverages meta-learning to separate known and unknown samples. Despite their promising performance, these methods focus more on known-class sampling and do not efficiently utilize unknown samples obtained during AL rounds.

To this end, we propose a novel AL framework designed to enhance the selection of informative samples from both known and unknown categories during the training process. Our approach aims to enhance the separation between known and unknown samples effectively. We achieve this by incorporating two distinct entropy scores into our framework. The first entropy score is computed using outputs from known classifiers. This score quantifies the uncertainty of samples with respect to the distribution of known classes. The second entropy score operates on a distance-based principle. Specifically, the unknown samples obtained from AL rounds are used to model the unknown distribution. Following this, the second score quantifies the uncertainty of samples with respect to the data distribution of unknown classes. Finally, utilizing the combined entropy score for all unlabeled samples we perform active sampling. Furthermore, to ensure diversity, we adaptively cluster and perform sampling on each cluster according to the AL budget. Extensive experiments show that our method outperforms existing state-of-the-art methods on CIFAR-10, CIFAR-100 and TinyImageNet datasets. 

The contributions of this paper are as follows.

\begin{itemize}
    \item We introduce an AL framework that leverages both known and unknown distributions to select informative samples during AL rounds.
    \item Specifically, we propose two entropy scores which separate the known and unknown samples for precise AL sampling. 
    \item Our experimental results show that the proposed method outperforms existing state-of-the-art methods on CIFAR-10, CIFAR-100, and TinyImageNet datasets.
\end{itemize}

\section{Related Work}
\begin{figure*}[t!]
    \begin{center}
        \includegraphics[width=0.80\linewidth]{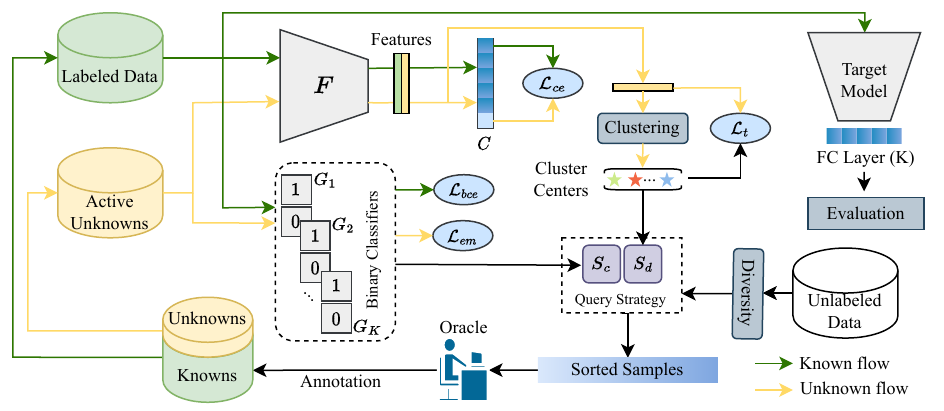}
    \end{center}
    \caption{
    Overview of the proposed method for open-set active learning. At each AL cycle, we begin by training $F$, $C$, $\left\{G_i\right\}_{i=1}^K$ via minimizing Eq. \ref{total_loss} on labeled data and active unknown data. Two distinct entropy scores are then computed to establish the query strategy $S$ as given in Eq. \ref{eq:uncertaity}. Next, we cluster the unlabeled samples into $K$ clusters and select the $\frac{b}{K}$ samples with the lowest $S$ values for annotation, where $b$ is the per-cycle annotation budget. Labeled and Active Unknown datasets are updated based on the annotated samples, and the target model is trained using the updated labeled dataset.
    }
    \label{fig:framework} 
\end{figure*}

\textbf{Active Learning (AL).} Active learning aims at maximizing the performance improvement of a model by choosing the most beneficial samples from a set of unlabeled data, labeling them, and incorporating them into the supervised training process. Uncertainty-based AL approaches attempt to select the samples that the model is most uncertain about via various uncertainty measures, such as entropy \cite{NIPS2013_b6f0479a}, mutual information \cite{kirsch2019batchbald} and confidence margin \cite{balcan2007margin}. On the other hand, diversity-based approaches \cite{sener2017active, xu2003representative, nguyen2004active} cluster the unlabeled samples and select representative samples from each cluster to better model the underlying distribution of unlabeled data. Query-by-Committee methods \cite{seung1992query, hino2022active} employ a measure of disagreement between an ensemble of models as the sample selection criterion. Recently, some methods have achieved enhanced AL performance by combining multiple sample selection criteria \cite{ash2019deep, wei2015submodularity, parvaneh2022active}. For example, \cite{ash2019deep} a combination of diversity and uncertainty is employed to achieve improved performance, using the model's gradient magnitude as a measure of uncertainty. However, while these standard AL methods excel in typical AL scenarios, they cannot perform as effectively in the open-set setting with a class distribution mismatch between labeled and unlabeled data.
\\
\noindent
\textbf{Open-set Recognition (OSR).} The problem of open-set recognition was first formulated in \cite{scheirer2012toward} and has gained significant traction in recent years. In \cite{bendale2016towards}, the authors introduce a method called OpenMax that trains a model with an extra class denoting the probability that a sample belongs to an open-set class. They utilize Extreme Value Theory (EVT) to calibrate the network's output for better OSR performance. \cite{ge2017generative, Neal_2018_ECCV, moon2022difficulty} use Generative Adversarial Networks (GANs) \cite{goodfellow2014generative} to synthesize images that resemble open-set classes. These images are then used to model the open-set space. Another line of work studies reconstruction-based approaches \cite{oza2019c2ae, sun2020conditional, yoshihashi2019classification} for open-set recognition. For example, C2AE \cite{oza2019c2ae} trains class-conditioned auto-encoders and models the reconstruction errors using EVT. During inference, the minimum value of class-wise reconstruction errors is compared with a predefined threshold to detect open-set samples. \cite{lu2022pmal, chen2020learning, yang2020convolutional, shu2020p} design different prototype-based learning mechanisms for OSR. \cite{chen2020learning} proposes a strategy called \textit{reciprocal point learning} that learns reciprocal points to represent \textit{otherness} of each known class. It attempts to push known samples far from reciprocal points and bound the open space of known classes. \cite{safaei2023open, cheng2023unified} propose the use of one-versus-all classifiers to enhance OSR performance. In general, the OSR task differs from the task of open-set AL in two critical aspects. First, in OSR, the entire set of known classes is labeled and accessible to the model during training, whereas in open-set AL, only a few labeled samples are available initially. Second, OSR algorithms lack access to the real unknown data during training, while open-set AL approaches must be specifically designed to fully utilize the knowledge obtained from unknown samples gathered in later rounds of AL. These differences highlight the importance of devising approaches tailored to the challenges posed by open-set AL.
\\
\noindent
\textbf{Open-set Active Learning.}
Recently, some approaches have studied the AL problem in the presence of open-set classes \cite{kothawade2021similar, du2021contrastive, park2022meta, ning2022active}. MQNet \cite{park2022meta} addresses the purity-informativeness trade-off in open-set AL by training an MLP that receives one open-set score and one AL score as input and outputs a balanced meta-score for sample selection. LfOSA \cite{ning2022active} attempts to construct a pure query set of known samples by modeling the maximum activation values of labeled data through class-wise GMMs and rejecting samples with lower probabilities as unknowns. However, as previously mentioned, many of these approaches do not fully utilize the availability of actual unknown data queried in AL rounds.

\section{Methodology}
In this section, we first present the problem of open-set AL, and then elaborate on the proposed approach in detail.
\\
\noindent
\textbf{Problem Formulation}. In open-set AL, we consider active learning for a $K$-way classification problem in an open-set setting, where $K$ denotes the number of classes of interest (known classes). In this setting, we are initially given a small labeled dataset $\mathcal{D}_{L}=\left\{ \left( x_{i}^{L},y_{i}^{L} \right) \right\}_{i=1}^{N_{L}}$ of \textit{known} samples that belong to the label space $\mathcal{K}=\left\{ j \right\}_{j=1}^{K}$, and a large pool of unlabeled data $\mathcal{D}_{U}=\left\{ x_{i}^{U} \right\}_{i=1}^{N_{U}}$ ($N_{L} \ll N_{U}$) which contains a mixture of \textit{known} and \textit{unknown} samples, where \textit{unknown} samples belong to the label space $\mathcal{U}$, and $\mathcal{K} \cap \mathcal{U}=\emptyset$ . 

At each AL cycle, the query strategy selects a batch of $b$ samples $X^{active}$ from the unlabeled data, and their labels are queried from an oracle. $X^{active}$ consists of \textit{known} queried samples $X^{k}$ and \textit{unknown} queried samples $X^{u}$. The samples in $X^{u}$ are labeled as the \textit{open-set} class (class $0$) by the oracle and referred to as \textit{active unknowns} in this paper. We add $X^{u}$ samples to the set of active unknowns $\mathcal{D}_{AU}$ and update the labeled dataset as $\mathcal{D}_L = \mathcal{D}_L \cup X^k$. The updated $\mathcal{D}_L$ is utilized to enhance the performance of a target model $T(\cdot)$ for an intended classification task.

\begin{algorithm}[tb]
	\caption{Our Proposed Algorithm for Open-set AL}
	\label{alg:LfOSA}
	\begin{algorithmic}[1]
		\STATE \textbf{Input:}
		\STATE \quad Labeled data $\mathcal{D}_L$, unlabeled data $\mathcal{D}_{U}$, number of AL 
  
  \quad cycles $R$, known categories $K$, per-cycle budget $b$, 
  
  \quad models $F$, $C$, $T$, and $\left\{G_i\right\}_{i=1}^K$
		\STATE \textbf{Process:}
		\STATE \quad $\mathcal{D}_{AU} \gets \emptyset \quad \quad $ \emph{{\# Initial active unknowns}}
            \STATE \quad Update models $F$, $C$, and $\left\{G_i\right\}_{i=1}^K$ by minimizing 
            
            \quad $\mathcal{L}_{total}$ in Eq. \ref{total_loss}
		
		\STATE \quad \textbf{for} $c=0,1,...,R-1$ \textbf{do}
		  \STATE \quad \quad $\forall x \in D_{U}$, $S_d(x)\gets 0$ \quad \quad \emph{{\# Initialization}}
		
		\STATE \quad \quad \textbf{if} $\mathcal{D}_{AU} \neq \emptyset$ \textbf{do}
            \STATE \quad \quad \quad Cluster the features of $\mathcal{D}_{AU}$ into $K$ clusters
            \STATE \quad \quad \quad For cluster $i$, compute the center $\textbf{c}_i$ using Eq. \ref{eq:cluster_centers}
            \STATE \quad \quad \quad $\forall x \in D_{U}$, compute $S_d(x)$ via Eq. \ref{ent2}
            \STATE \quad \quad \textbf{end if}
		
		\STATE \quad \quad $\forall x \in D_{U}$, compute $S_c(x)$ via Eq. \ref{ent1}
            \STATE \quad \quad Cluster the features of $\mathcal{D}_{U}$ into $K$ clusters, 
            
            \quad \quad $\left\{C_1, C_2, ..., C_K\right\}$ \quad \quad \emph{{\# Diversity}}
            
            \STATE \quad \quad \textbf{for} $j = 1,2,...,K$ \textbf{do}
            \STATE \quad \quad \quad $\mathcal{S}_j \gets \left\{ S_c(x) - S_d(x)|\forall x \in C_j\right\}$ \emph{{\# Uncertainty}}
		\STATE \quad \quad \quad $X_j \gets$ select the $\frac{b}{K}$ samples with the lowest 
        
        \quad \quad \quad values from $\mathcal{S}_j$, and annotate them
        \STATE \quad \quad \textbf{end}
        \STATE \quad \quad \emph{{\# All queries for the current cycle:}}
        \STATE \quad \quad $X^{active} \gets X_1 \cup X_2 \cup ... \cup X_K$ 
        \STATE \quad \quad \emph{{\# Knowns and active unknowns:}}
        \STATE \quad \quad Obtain $X^k$ and $X^u$ 
        \STATE \quad \quad $\mathcal{D}_{L} \gets \mathcal{D}_{L} \cup X^k$,  $\mathcal{D}_{AU} \gets \mathcal{D}_{AU} \cup X^{u}$,
            \STATE \quad \quad $\mathcal{D}_{U} \gets \mathcal{D}_{U} / X^{active}$ \quad \emph{{\# Update datasets}}
            \STATE \quad \quad Update the target model $T$ via minimizing the 
            
            \quad \quad cross-entropy loss on $D_L$
            \STATE \quad \textbf{end}
            
	\end{algorithmic}
 \label{alg:algorithm}
 
\end{algorithm}
\noindent
\textbf{Overview.} Our AL framework utilizes two entropy scores that effectively differentiate between known and unknown samples, making them suitable for selecting the most valid samples for annotation. First, the \textit{Closed-set Entropy} is calculated based on the outputs of $K$ class-aware binary classifiers (BC) trained on $\mathcal{D}_L$. This entropy quantifies the uncertainty of a sample with respect to the distributions of the known classes, which tends to be low for known samples and high for unknown samples. Second, the \textit{Distance-based Entropy} is utilized to prioritize the selection of samples that stand apart from distributions of unknown classes.  To compute this entropy for a given sample, we start by clustering the Convolutional Neural Network (CNN) based features of $\mathcal{D}_{AU}$ samples and determining the cluster centers. The distances between the sample and these cluster centers are then used to measure the entropy of the samples. Fig. \ref{fig:framework} provides an overview of our approach.

\subsection{Training for Closed-set Entropy Scoring}
To quantify closed-set entropy score ($S_c$), we employ (1) a CNN-based feature extractor $F(\cdot)$, (2) $K$ class-aware binary classifiers $G_i(\cdot)$, $i \in \left\{ 1,2,...,K \right\}$, and (3) a fully-connected layer $C(\cdot)$ that produces a probability vector $\in \mathbb{R}^{K+1}$ for ($K+1$)-way classification on $\mathcal{D}_L \cup \mathcal{D}_{AU}$. The parameters of $F$ and $C$ are updated using a standard cross-entropy loss ($\mathcal{L}_{ce}$) on $\mathcal{D}_L \cup \mathcal{D}_{AU}$.
\\
\noindent
\textbf{Training Binary Classifiers.}
We train each $G_i$ with the samples in the $i$-th known class as positives, and the remaining known samples as negatives. For a given image $\textbf{x}$, let  $\textbf{f} = F\left( \textbf{x} \right)$ denote the extracted features of $\textbf{x}$, and $p^i=\sigma\left( G_i\left( \textbf{f} \right) \right)$ denote the probability of \textbf{x} being categorized as positive class by $G_i$, where $\sigma$ is the Softmax operator. The loss function for training $G_i$'s is as follows: 
\begin{equation}
    \mathcal{L}_{bce} = \frac{1}{n_l}\sum_{(\textbf{x}_i,y_i) \in \mathcal{D}_L} -\log ({p^{y_{i}}})-\min _{j \neq y_{i}} \log (1 - {p^{j}}),
\label{modified_bce}
\end{equation}
where $n_l$ is the number of known samples in the batch. This loss is a modified version of binary cross-entropy (BCE) \cite{saito2021ovanet} that only updates the positive and the nearest negative decision boundaries for each sample. This is to mitigate the bias of a BC towards the negative class which includes lots of samples.
\\
\noindent
\textbf{Closed-set Entropy.} For a given sample $\textbf{x}$, we define closed-set entropy score as follows:
\begin{equation}
    S_{c}(\textbf{x})=\frac{1}{K\cdot\log(2)}\sum_{i=1}^{K} H_i(\textbf{x}),
    \label{ent1}
\end{equation}
where $H_i(\textbf{x})$ denotes the entropy of $G_i$ given as:
\begin{equation}
   H_i(\textbf{x}) = -p^{i} \cdot \log ({p^{i}}) -\left( 1-p^{i} \right) \cdot \log ({1-p^{i}}).
\label{entropy}
\end{equation}
$S_{c}$ measures the average normalized entropy of BC's.
\\
\noindent
\textbf{High $S_c$ on Unknown Samples.} 
While training a BC via Eq. (\ref{modified_bce}) obviously ensures its low entropy for known samples, it does not necessarily guarantee a high entropy for all unknown samples. To ensure high $S_c$ for unknown samples, we minimize the following objective on $\mathcal{D}_{AU}$:
\begin{equation}
    \mathcal{L}_{em} = \frac{1}{K\cdot n_{au}}\sum_{\textbf{x} \in \mathcal{D}_{AU}} \sum_{i=1}^{K}-\frac{1}{2}\log ({p^{i}}) -\frac{1}{2}\log ({1-p^{i}}),
\end{equation}
where $n_{au}$ is the number of active unknown samples in the batch. This loss encourages uniform probability outputs $p = [\frac{1}{2},\frac{1}{2}]$ for unknown samples.  
\\
\noindent \textbf{Property 1.} \textit{Minimizing $L_{em}$ is equivalent to maximizing the entropy of each $G_i$.}
\\
\noindent
\textbf{\textit{Proof}.} We have $p^i + (1-p^i) = 1$, and $p^i \in (0,1)$. By applying Jensen's inequality for concave functions, we obtain $H_i(\textbf{x}) \leq \log (2)$ and $\mathcal{L}_{em} \geq \log (2) $, where the equality happens \textit{iff} $p^i = (1-p^i) = \frac{1}{2}$.

\subsection{Training for Distance-based Entropy Scoring}
$S_c$ alone is insufficient for open-set active sampling, as it can be misled by unknown samples in close proximity to a known category. Hence, we employ a distance-based entropy score $S_d$ to achieve higher precision in selecting known samples. Typically, an unknown sample lies near the distribution of its ground-truth category while being distant from other categories, and hence it exhibits a low $S_d$. Conversely, a known sample remains distant from all unknown categories, resulting in a high $S_d$.  
\\
\noindent
\textbf{Distance-based Entropy.} We leverage $\mathcal{D}_{AU}$ samples for computing $S_d$. Having no access to their precise category labels, we first cluster these samples using the FINCH clustering algorithm \cite{sarfraz2019efficient}. We fix the number of clusters to $K$. Denoting the obtained cluster labels for $\mathcal{D}_{AU}$ samples as $\left\{ \hat{y}_i \right\}_{i=1}^{N_{AU}}$, we then compute the center of each cluster as follows: 
\begin{equation}
    \mathbf{c}_i = \frac{\sum_{(\textbf{x},\hat{y})\in \mathcal{D}_{AU}} \mathbb{I}\{\hat{y} = i\} \cdot F(\textbf{x})}{\sum_{(\textbf{x},\hat{y})\in \mathcal{D}_{AU}}\mathbb{I}\{\hat{y} = i\}} ,
\label{eq:cluster_centers}
\end{equation}
where $\mathbb{I}\{\cdot\}$ is the indicator function, and $N_{AU} = \left| \mathcal{D}_{AU}\right|$. Finally, $S_d$ is defined as:
\begin{equation}
\begin{aligned}
        S_{d}(\textbf{x})=\frac{-1}{\log(K)}\sum_{i=1}^{K} q_i(\textbf{x}) \cdot \log (q_i(\textbf{x})),
     \\ q_i (\textbf{x})= \frac{e^{-\left\| F\left ( \textbf{x} \right ) - \textbf{c}_i\right\|/T}}{\sum_{j=1}^{K}e^{-\left\| F\left ( \textbf{x} \right ) - \textbf{c}_j\right\|/T}},
\end{aligned}
\label{ent2}
\end{equation}
where $q_i (\textbf{x})$ is the probability of a sample $\textbf{x}$ belonging to the $i$-th cluster, and $T$ is a temperature. Basically, we calculate the distances of a sample from cluster centers, form the probability vector $[q_1(\textbf{x}),q_2(\textbf{x}),...,q_K(\textbf{x})]$, and compute its normalized entropy. We chose FINCH over K-means \cite{macqueen1967some} due to its superior speed and efficiency.
\\
\noindent
\textbf{Low $S_d$ on Unknown Samples.} Using cross-entropy loss for training $F(\cdot)$ generally ensures high distance of known samples from $\textbf{c}_i$'s in the feature space, resulting in high $S_{d}$ values. However, low $S_{d}$ values for unknown samples cannot be guaranteed since cross-entropy loss treats all active unknowns as one open-set class. Hence, we further regularize the features to impose more compactness within each cluster while maintaining a larger margin between different clusters. Specifically, for a sample $\textbf{x} \in \mathcal{D}_{AU}$ with cluster label $\hat{y}$, we utilize Tuplet loss \cite{sohn2016improved, miller2021class} to enforce a low distance between $F(\textbf{x})$ and the cluster center $\textbf{c}_{\hat{y}}$, and a large margin between the distance to $\textbf{c}_{\hat{y}}$ and the distance to $\textbf{c}_{j\neq \hat{y}}$. The loss can be written as follows:
\begin{equation}    
   \mathcal{L}_{t} = \frac{1}{ n_{au}}\sum_{\textbf{x} \in \mathcal{D}_{AU}} \log\Big(1 + \sum_{j \neq \hat{y}}^{K} e^{D_{\hat{y}} - D_j}\Big)+\beta  D_{\hat{y}},
\end{equation}
where $D_i=\left\| F\left ( \textbf{x} \right ) - \textbf{c}_i\right\|$.
\subsection{Query Strategy}
\begin{figure*}[t!]
    \begin{center}
        \includegraphics[width=1\linewidth]{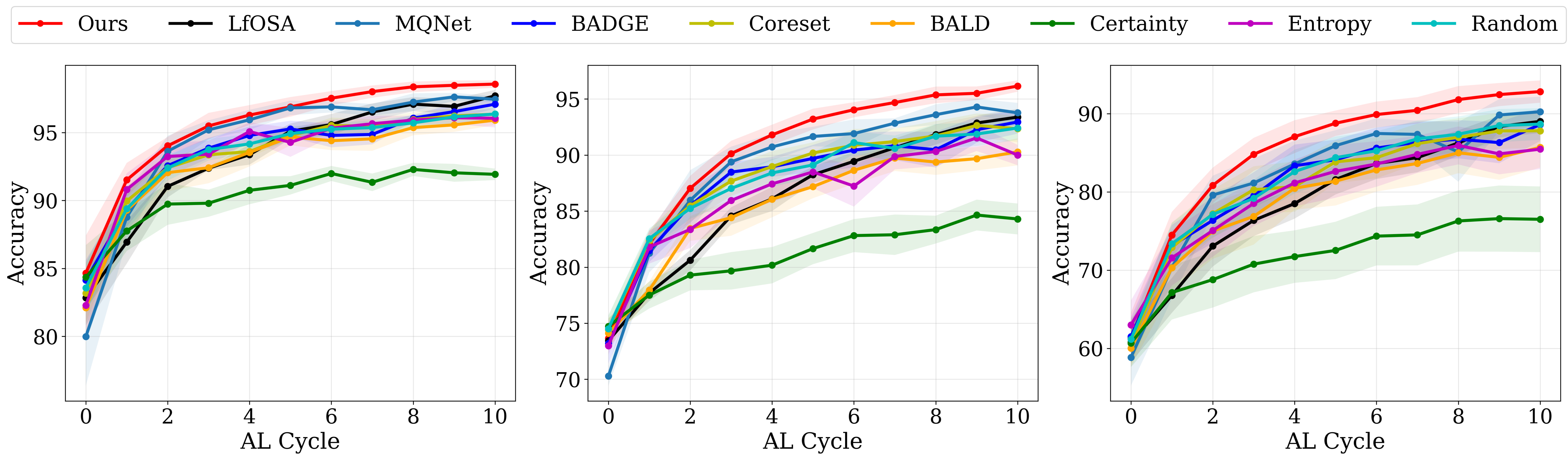}
    \end{center}
    \vskip -15.0pt
    \caption{
    Classification accuracy comparison on CIFAR-10 (mismatch ratios from left to right: $20\%$, $30\%$, and $40\%$).
    }
    \label{fig:cifar10_results} 
    \vskip -10.0pt
\end{figure*}

\begin{figure*}[t!]
    \begin{center}
        \includegraphics[width=1\linewidth]{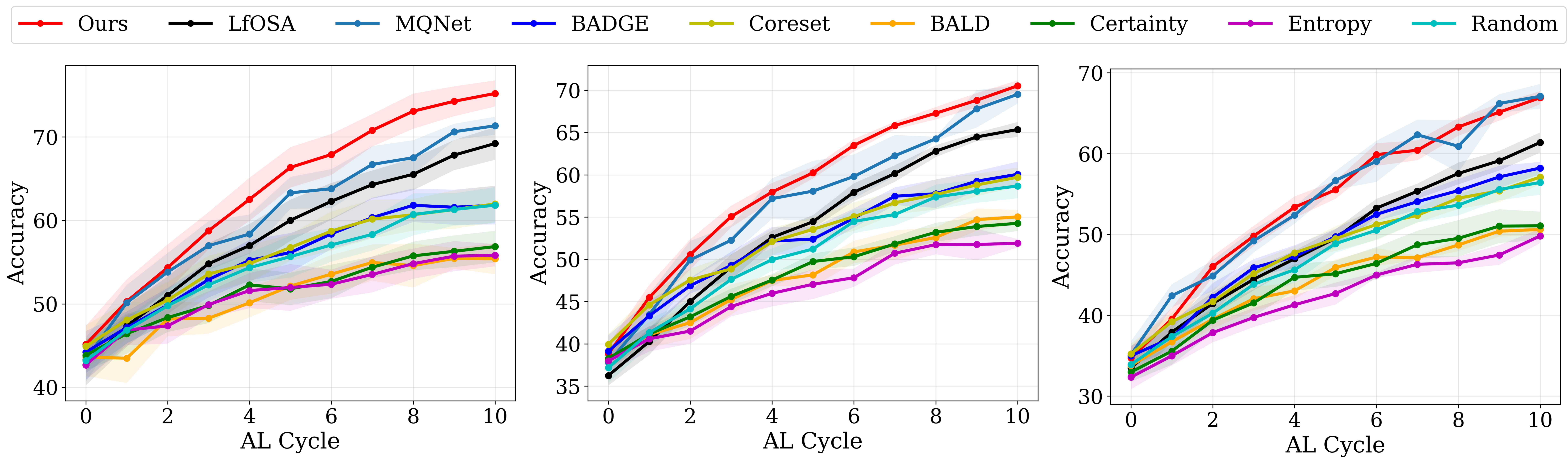}
    \end{center}
    \vskip -15.0pt
    \caption{
    Classification accuracy comparison on CIFAR-100 (mismatch ratios from left to right: $20\%$, $30\%$, and $40\%$).
    }
    \label{fig:cifar100_results} 
    \vskip -10.0pt
\end{figure*}

\begin{figure*}[t!]
    \begin{center}
        \includegraphics[width=1\linewidth]{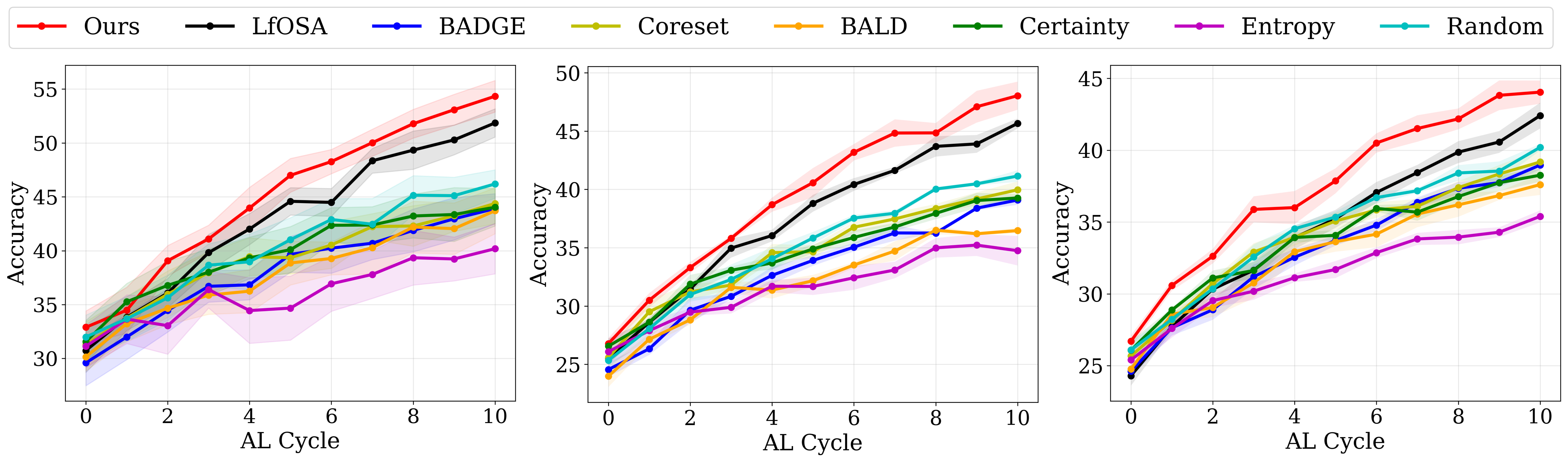}
    \end{center}
    \vskip -15.0pt
    \caption{
    Classification accuracy comparison on TinyImageNet (mismatch ratios from left to right: $20\%$, $30\%$, and $40\%$).
    }
    \label{fig:tin_results} 
    \vskip -10.0pt
\end{figure*}
As described in previous sections, we formulate our query strategy by combining Eq. \ref{ent1} and Eq. \ref{ent2} as 
\begin{equation}
    S(\textbf{x}) = S_{c}(\textbf{x}) - S_{d}(\textbf{x}).
    \label{eq:uncertaity}
\end{equation}
$S(\textbf{x})$ score estimates the uncertainty of a sample with respect to the data distributions of both known ($S_{c}$) and unknown ($S_{d}$) categories. A known sample remains near its corresponding known category ($S_{c}\downarrow$) and distinct from all unknown categories ($S_{d}\uparrow$), while the opposite holds for an unknown sample ($S_{d}\downarrow$ and $S_{c}\uparrow$). As a result, $S(\textbf{x})$ can effectively separate known samples from unknown ones. 

Furthermore, to select informative samples, we propose to query from different regions in the feature space to minimize the redundancy of the selected samples. To be specific, our query strategy is as follows. We first consider the unlabeled samples which are classified as one of the known categories by $C(\cdot)$ and cluster them into $K$ clusters using the FINCH algorithm. Then, within each cluster, we sort the samples based on the $S(\textbf{x})$ score and select the first $\frac{b}{K}$ samples with the lowest scores for annotation, where $b$ denotes the per-cycle annotation budget (see Algorithm \ref{alg:algorithm}). The importance of each component of our query strategy is further studied in ablation studies.
\subsection{Overall Loss} Our proposed method is trained in an end-to-end manner by minimizing the following total objective:
\begin{equation}
    \mathcal{L}_{total} =
    \begin{cases}
        \mathcal{L}_{ce}+\mathcal{L}_{bce} & \text{if $\mathcal{D}_{AU}=\emptyset$}\\
        \mathcal{L}_{ce}+\mathcal{L}_{bce}+\mathcal{L}_{em}+{\lambda}\mathcal{L}_{t} & \text{if $\mathcal{D}_{AU}\neq \emptyset$}\\
       
    \end{cases},
\label{total_loss}
\end{equation}
where we minimize $\mathcal{L}_{bce}$ on $\mathcal{D}_L$, $\mathcal{L}_{em}$ and $\mathcal{L}_{t}$ on $\mathcal{D}_{AU}$, and $\mathcal{L}_{ce}$ on $\mathcal{D}_{L}\cup\mathcal{D}_{AU}$. Note we do not consider $\mathcal{L}_{em}$ and $\mathcal{L}_{t}$ in the total objective before the first AL cycle since $\mathcal{D}_{AU}=\emptyset$.
\subsection{Training the Target Model}
After querying the samples at each AL cycle, a target model is trained on the updated $\mathcal{D}_L$ dataset, using the standard cross-entropy loss. The performance of this model in $K$-way classification is utilized for our evaluations.
\section{Experiments}

We perform extensive experiments on the CIFAR-10, CIFAR-100 \cite{krizhevsky2009learning}, and TinyImageNet \cite{yao2015tiny} datasets to demonstrate the effectiveness of our approach. 
\\
\noindent
\textbf{Datasets.} The CIFAR-10 and CIFAR-100 datasets each contain 50000 images for training and 10000 images for testing, while they consist of 10 and 100 categories, respectively. TinyImageNet is a large-scale dataset containing 100000 training images and 20000 testing images in 200 categories.
\\
\noindent
\textbf{Experimental Setting.} 
For each dataset, we consider some randomly chosen classes to be the \textit{knowns} and the remaining classes to be the \textit{unknowns} using a mismatch ratio. The mismatch ratio is defined as $\frac{\left| \mathcal{K} \right|}{\left| \mathcal{K} \right|+\left| \mathcal{U} \right|}$, where $\left| \mathcal{K} \right|$ is the number of known classes and $\left| \mathcal{U} \right|$ is the number of unknown classes. For CIFAR-10, CIFAR-100, and TinyImageNet, we initialize the labeled dataset by randomly sampling $1\%$, $8\%$, and $8\%$ of the samples from known classes, respectively. In all of our experiments, we perform $10$ cycles of active sampling, and $1500$ samples are queried for annotation in each cycle. For fair experimental results, each experiment is conducted four times with varying known/unknown class splits across all the compared methods. The average results from these runs are then reported.
\\
\noindent
\textbf{Implementation Details.}
In all experiments, we train a ResNet18 \cite{he2016deep} as our backbone network and one-layer fully-connected networks as binary classifiers. In each AL cycle, we train models for 300 epochs via SGD optimizer \cite{ruder2016overview} with an initial learning rate of 0.01, a momentum of 0.9, and a weight decay of 0.005. The learning rate is decayed by 0.5 every 60 epochs. The batch size is set to 128 for all experiments. We generally set the values of both $\beta$ and $\lambda$ to 0.1. We utilize PyTorch \cite{paszke2019pytorch} to implement our method and an NVIDIA A5000 GPU to run each experiment. We do not use any pre-trained model.
\\
\noindent
\textbf{Baselines.} We compare our method with the following AL and open-set AL approaches, namely Random, Entropy, Certainty, BALD \cite{tran2019bayesian}, Coreset \cite{sener2017active}, BADGE \cite{ash2019deep}, MQNet \cite{park2022meta}, and LfOSA \cite{ning2022active}, from which LfOSA and MQNet are the SOTA methods for open-set AL. Random sampling randomly selects unlabeled samples for annotation. Entropy \cite{Wang2014ANA} selects samples based on the model's predictive entropy. Certainty \cite{NIPS2013_b6f0479a} chooses samples with high confidence in their predictions. BALD \cite{tran2019bayesian} uses the uncertainty Bayesian CNN provides to select informative samples. Coreset \cite{sener2017active} utilizes the concept of core-set selection to choose diverse samples. BADGE \cite{ash2019deep} employs both uncertainty and diversity for active sampling. MQNet \cite{park2022meta} employs a meta-learning approach to combine an open-set score and an AL criterion for sample selection. LfOSA \cite{ning2022active} uses GMM to reject open-set samples and achieve higher known purity in the query set. 
\subsection{Main Results}
\begin{figure*}[t!]
    \begin{center}
        \includegraphics[width=1\linewidth]{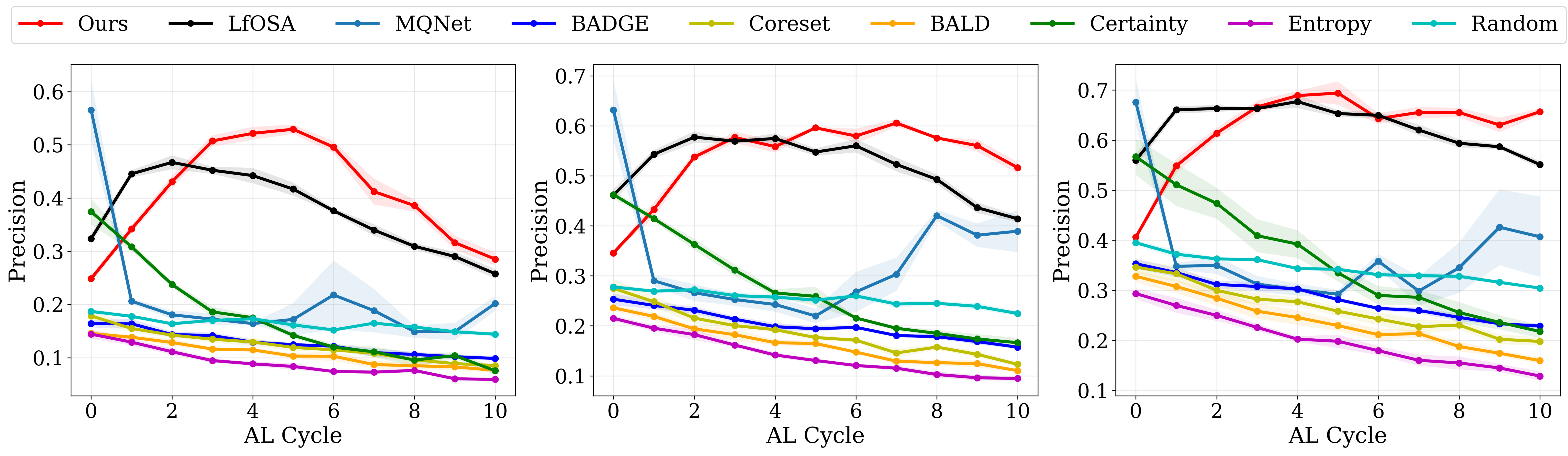}
    \end{center}
    \vskip -15.0pt
    \caption{
    Precision results on CIFAR-100 (mismatch ratios from left to right: $20\%$, $30\%$, and $40\%$).
    }
    \label{fig:precision} 
    \vskip -10.0pt
\end{figure*}

\begin{figure*}[t!]
    \begin{center}
        \includegraphics[width=1\linewidth]{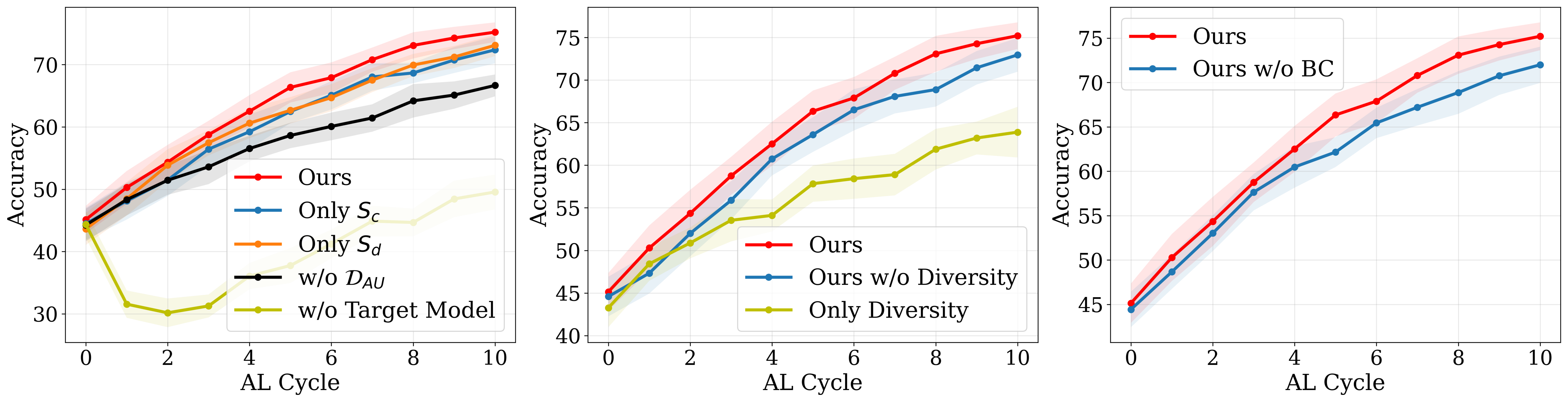}
    \end{center}
    \vskip -5.0pt
    \caption{
    Ablation study on CIFAR-100 with the mismatch ratio of $20\%$
    }
    \label{fig:ablation} 
    \vskip -5.0pt
\end{figure*}
\noindent
\textbf{Classification Results.} Fig. \ref{fig:cifar10_results}, \ref{fig:cifar100_results}, and \ref{fig:tin_results} show the classification results corresponding to various methods on CIFAR-10, CIFAR-100, and TinyImageNet, respectively. It can be seen that the proposed approach outperforms other baselines nearly across all datasets and mismatch ratios. Our method can effectively identify known samples within unlabeled data by utilizing the proposed entropy scores in the query strategy. As a result, it shows excellent performance in challenging scenarios with high unknown ratios. Specifically, our method outperforms recent open-set AL methods MQNet and LfOSA by margins of $3.88\%$ and $6.00\%$, respectively, on CIFAR-100 with the mismatch ratio of $20\%$. As the mismatch ratio increases, we observe a drop in the performance gap between LfOSA and standard AL methods. This is because LfOSA mainly relies on the purity of selected samples, which becomes less effective in high mismatch ratios, where there is an abundance of unlabeled known data. In contrast, our approach maintains a large performance margin compared to the standard AL methods by sampling from clusters of unlabeled data to ensure diversity. 
\\
\noindent
\textbf{Precision Results.} The precision of different AL methods in selecting known samples is shown in Fig. \ref{fig:precision}. As shown in this figure, LfOSA maintains a high precision by focusing on selecting as many known samples as possible which can lead to sampling uninformative samples in low unknown ratio scenarios. Conversely, the precision of MQNet declines rapidly after the first few cycles which does not yield optimal results when the unknown ratio is high due to prioritizing informativeness over purity. However, our approach strikes a balance between these two methods. It maintains high precision across AL cycles by employing two distinct entropy scores and simultaneously selects diverse samples through sampling from clusters. This shows the effectiveness of our method in both high and low unknown ratio settings.
 
\subsection{Ablation Study}

In this section, we conduct the ablation study on CIFAR-100 with the mismatch ratio of $20\%$ to show the effectiveness of each component within our framework. Fig. \ref{fig:ablation} (left) studies the following cases:
\\
\noindent
\textbf{Only} $S_{c}$. It indicates using only the $S_c$ in the query strategy. Accordingly, we do not utilize $\mathcal{L}_{t}$ in this setting.
\\
\noindent
\textbf{Only} $S_{d}$. It uses only $S_d$ in the query strategy. Accordingly, we do not leverage $\mathcal{L}_{em}$ in this setting. It can be observed that removing each entropy score leads to reduced accuracy performance across all AL cycles. This shows the effectiveness of combining these two entropy scores for the query strategy in our framework.
\\
\noindent
\textbf{w/o} $\mathcal{D}_{AU}$. It denotes we do not utilize $\mathcal{D}_{AU}$ in any part of our method training. We can observe the importance of training with active unknown samples to achieve satisfactory open-set AL performance. 
\\
\noindent
\textbf{w/o Target Model}. It indicates the utilization of the trained feature extractor $F$ and the classifier $C$ for the final evaluation on the testing data, as opposed to training a separate target model on $\mathcal{D}_L$. The observed performance drop emphasizes the need for a separate target model for evaluations.

In Fig. \ref{fig:ablation} (middle), we evaluate the importance of diversity in our proposed query strategy as follows:
\\
\noindent
\textbf{w/o Diversity}. In this experiment, clustering is not utilized in the query strategy. Instead, we choose the samples with the lowest $S(\textbf{x})$ scores from all unlabeled data globally, rather than from each cluster. We see the performance of the last AL cycle drops by $2.25\%$ compared to our approach.
     \\
\noindent
\textbf{Only Diversity}. In this experiment, we randomly select samples from each cluster, rather than sorting them by the $S(\textbf{x})$ score. The performance decreases by $11.33\%$ indicating that diversity sampling alone is not effective for selecting informative and valid samples. 

In Fig. \ref{fig:ablation} (right) we study the role of binary classifiers $\left\{G_i\right\}_{i=1}^K$ in our framework:
\\
\noindent
\textbf{w/o BC}. In this experiment, we remove the BC block from our framework. Not using $\left\{G_i\right\}_{i=1}^K$ to form the $S_c$ score, we utilize the first $K$ logit outputs of $C$ to calculate the closed-set entropy. The accuracy declines by a margin of $3.21\%$ in this setting which shows the effectiveness of utilizing BC's in our framework. 
\definecolor{lightgray}{RGB}{220,220,220}

\section{Conclusion}
In this paper, we propose a novel framework for addressing the problem of open-set active learning where we leverage both known and unknown class distribution. Specifically, our approach includes a closed-set entropy score that quantifies the uncertainty of a sample with respect to distributions of known categories and a distance-based entropy that measures uncertainty regarding distributions of unknown categories. By utilizing these entropy scores, we effectively separate the known and unknown samples, and followed by clustering, we select the most informative samples. We conducted extensive experiments on CIFAR-10, CIFAR-100, and TinyImageNet, showing our proposed approach's effectiveness in both high and low open-set noise ratio scenarios. 

\section* {Acknowledgements}
Research was sponsored by the Army Research Laboratory and was accomplished under
Cooperative Agreement Number W911NF-23-2-0008. The views and conclusions contained in this document are those of the authors and should not be interpreted as representing the official policies, either expressed or implied, of the Army Research Laboratory or the U.S. Government. The U.S. Government is authorized to reproduce and distribute reprints for Government purposes notwithstanding any copyright notation herein.
\bibliography{aaai24}

\end{document}